\documentclass[10pt,conference,compsoc]{IEEEtran}

\usepackage{cite}
\usepackage{multirow}
\usepackage{subfigure}
\usepackage{amssymb}
\usepackage{xspace}
\usepackage{epsf}
\usepackage{setspace}
\usepackage{caption}
\usepackage{comment}
\usepackage{graphicx}
\usepackage{amsmath}

\usepackage[ruled,vlined,linesnumbered,ruled]{algorithm2e}
\usepackage{balance}
\usepackage{url,epsfig, array}
\usepackage{amsmath}
\usepackage{epstopdf}
\usepackage{cases}
\usepackage{bm}
\usepackage{color}
\usepackage{tikz}
\usepackage{threeparttable}

\usepackage{diagbox}

\newcommand*{\circled}[1]{\lower.7ex\hbox{\tikz\draw (0pt, 0pt)%
		circle (.5em) node {\makebox[1em][c]{\small #1}};}}

\def\ie{\textit{i.e.}\xspace}

\def\eg{\textit{e.g.}\xspace}

\begin{document}
	\title{An Efficient NAS-based Approach for Handling Imbalanced Datasets}
	\author{
		% Yang Xu,~\IEEEmembership{Member,~IEEE,}~
		% Yunming Liao,~
		% $^*$Hongli Xu,~\IEEEmembership{Member,~IEEE,}~
		% Zhenguo Ma,\\
		% Lun Wang,~
		Zhiwei Yao \\
        School of Computer Science and Technology \\
        University of Science and Technology of China \\ 
		% % Jianchun Liu,~\IEEEmembership{Member,~IEEE,ACM,}
  %       Yang Xu,~\IEEEmembership{Member,~IEEE,}~
		% Hongli Xu,~\IEEEmembership{Member,~IEEE,}~
  %       Yunming Liao,~
		% Lun Wang~
		% Chen Qian,~\IEEEmembership{Senior Member,~IEEE,}~
        % Yunming Liao ~ 
%		Yangming Zhao,~\IEEEmembership{Fellow,~IEEE}
%		Yangming Zhao, ~\IEEEmembership{}
		% Yangming Zhao  ~\IEEEmembership{}
		% \IEEEcompsocitemizethanks{
		% 	\IEEEcompsocthanksitem Z. Yao, Y. Xu, H. Xu, Y. Liao, and L. Wang are with the School of Computer Science and Technology, University of Science and Technology of China, Hefei, Anhui, China, 230027, and also with Suzhou Institute for Advanced Research, University of Science and Technology of China, Suzhou, Jiangsu, China, 215123. \protect E-mails: zhiweiyao@mail.ustc.edu.cn; xuyangcs@ustc.edu.cn;  xuhongli@ustc.edu.cn; liaoyun@mail.ustc.edu.cn; wanglun0@mail.ustc.edu.cn. 
		% 	% \IEEEcompsocthanksitem C. Qian is with the Department of Computer Science and Engineering, Jack Baskin School of Engineering, University of California Santa Cruz, Santa Cruz, CA 95064 USA. E-mail: cqian12@ucsc.edu.
		% 	% qiao@buffalo.edu
		% 	% \IEEEcompsocthanksitem H. Xu is the corresponding author.
  %           \IEEEcompsocthanksitem Y. Xu is the corresponding author.
		% }
	}

	\markboth{IEEE Transactions on Mobile Computing, Vol., No., Nov. 2023}%
	{Shell \MakeLowercase{\textit{et al.}}: Bare Advanced Demo of IEEEtran.cls for Journals}
	
	\IEEEtitleabstractindextext{%
		\begin{abstract}
			Class imbalance is a common issue in real-world data distributions, negatively impacting the training of accurate classifiers. Traditional approaches to mitigate this problem fall into three main categories: class re-balancing, information transfer, and representation learning. In this paper, we introduce a novel approach to enhance performance on long-tailed datasets by optimizing the backbone architecture through neural architecture search (NAS). Our research shows that an architecture’s accuracy on a balanced dataset does not reliably predict its performance on imbalanced datasets. This necessitates a complete NAS run on long-tailed datasets, which can be computationally expensive.
To address this computational challenge, we focus on existing work, called IMB-NAS, which proposes efficiently adapting a NAS super-network trained on a balanced source dataset to an imbalanced target dataset.
A detailed description of the fundamental techniques for IMB-NAS is provided in this paper, including NAS and architecture transfer.
Among various adaptation strategies, we find that the most effective approach is to retrain the linear classification head with reweighted loss while keeping the backbone NAS super-network trained on the balanced source dataset frozen. 
Finally, we conducted a series of experiments on the imbalanced CIFAR dataset for performance evaluation. 
Our conclusions are the same as those proposed in the IMB-NAS paper.

		\end{abstract}
	}
	
	\maketitle
	\IEEEdisplaynontitleabstractindextext
	\IEEEpeerreviewmaketitle
	
	% \vspace{-2mm}
	\section{Introduction}\label{sec_introduction}
        
\begin{figure}[!t]
	\centering
	\includegraphics[width=0.9\columnwidth]{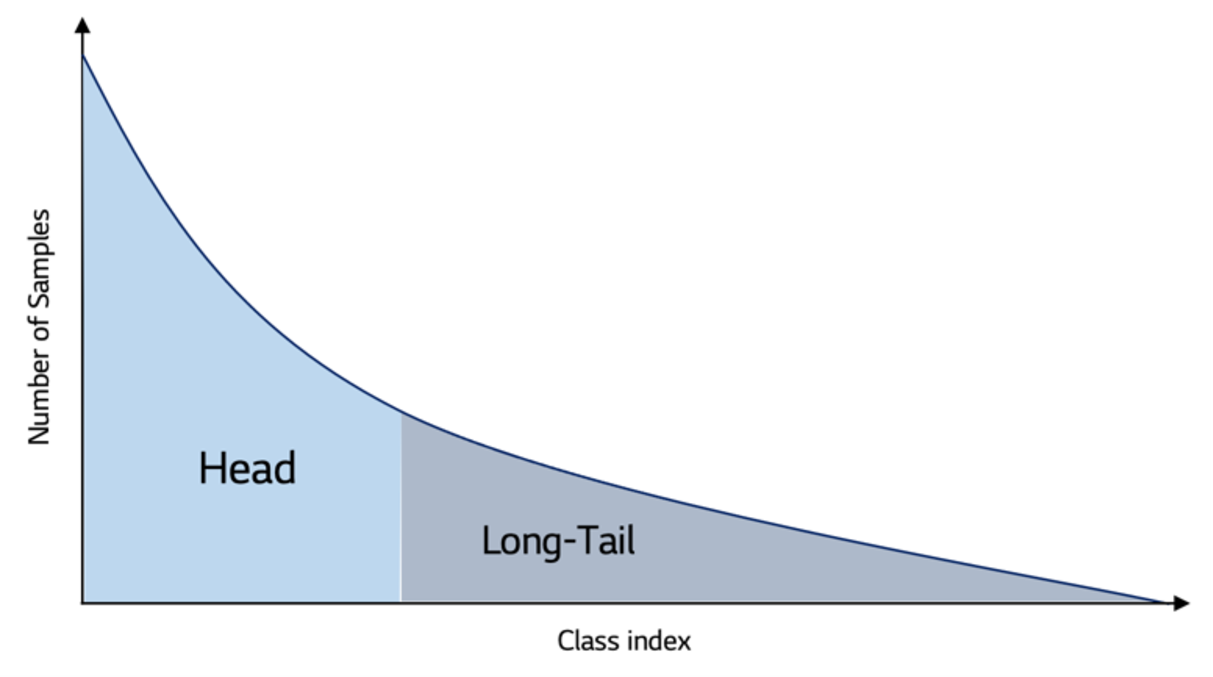}
	\caption{Long-tailed data distribution.}
	\label{fig: Long-tailed}
	% \vspace{-0.5cm}
\end{figure}

The natural world exhibits a long-tail data distribution, as shown in Fig.\ref{fig: Long-tailed}, where a small number of classes dominate the majority of data samples, while the remaining data is spread across numerous minority classes.
Much of the previous work \cite{kang2019decoupling, zhou2020bbn, duggal2021har} has concentrated on improving the accuracy of fixed backbone architectures like ResNet-32.
In contrast, our work aims to optimize the backbone architecture using neural architecture search (NAS). 
This is particularly important as current practices require neural architectures to be optimized for the size and latency constraints of small edge devices.

To enhance the backbone architecture, we leverage recent advancements in Neural Architecture Search (NAS) \cite{guo2020single}, which primarily focuses on datasets that are balanced across classes. 
This raises a critical question: Is an architecture optimized on a balanced dataset also optimal for imbalanced datasets? 
Obviously, when performing architecture search and model training on imbalanced datasets, the model is prone to bias towards the head classes with massive
samples. This bias results in significantly lower accuracy on the tail classes, which have only a few samples.
For example, Duggal \cite{duggal2022imb} conducted experiments where they trained identical architectures on datasets with varying distributions. Their findings revealed that the performance of the model varied substantially depending on the distribution of the data. 
Specifically, the architectures showed high accuracy on balanced datasets but struggled with imbalanced datasets.
This demonstrates the critical need for effective strategies to mitigate class imbalance during both the search and training phases of neural network development.

Executing a NAS procedure for each target dataset demands significant computational resources and rapidly becomes unfeasible when dealing with multiple target datasets.
To address this challenge,  we learn the scheme proposed in \cite{duggal2022imb}, which proposes a more efficient approach: adapting architectural rankings from balanced datasets to imbalanced ones. This approach leverages the strength of NAS while minimizing computational costs. Specifically, it focuses on reusing a NAS super-network trained on balanced data and adapting it to imbalanced data by retraining only the linear classification head. This strategy significantly reduces the computational burden as it involves training only a linear layer on top of the pre-trained super-network.
Extensive experiments in IMB-NAS\cite{duggal2022imb} reveal a key insight: the adaptation procedure is most influenced by the linear classification head trained on top of the backbone. This finding suggests that the backbone, once trained on a balanced dataset, can generalize well to imbalanced datasets with minimal additional training. 
Based on this insight, we implement this scheme over the imbalanced CIFAR dataset \cite{krizhevsky2009learning}, reuse a NAS super-network backbone trained on balanced CIFAR-10 and retrain only the classification head to adapt efficiently to imbalanced CIFAR-100. This method is highly efficient as it involves training only a linear layer on top of the pre-trained super-network.

The remainder of the paper is organized as follows.
Section 2 discusses some related work, which includes neural architecture search, long-tailed data learning and architecture transfer. 
In Section 3, we first introduce some preliminaries about the DARTS\cite{liu2018darts} and present a detailed description of the scheme in \cite{duggal2022imb}.
Section 4 presents the experimental setup and evaluation results, and
finally, the paper is concluded in Section 5.

        \section{Related Works}\label{sec:related}
        \subsection{Neural architecture search}
 NAS is a method for automating the design of neural network architectures, typically involving search space design, search strategy design, and performance estimation strategy.
Search space design creates a diverse range of possible architectures, such as cell-based spaces like NASNets \cite{zoph2018learning} and DARTS, or macro search spaces like those used in ShuffleNet \cite{zhang2018shufflenet} and MobileNet \cite{howard2017mobilenets} models. 
Search Strategy Design focuses on efficiently identifying high-performing architectures within the search space. Common strategies include reinforcement learning \cite{baker2016designing, zoph2018learning}, where RL agents iteratively propose and evaluate architectures, receiving rewards based on their performance.
Evolutionary algorithms \cite{real2017large, duggal2021compatibility} apply principles such as mutation and selection to evolve a population of architectures, exploring a broad range of designs.
Gradient-based methods, such as those used in DARTS \cite{liu2018darts}, optimize architectures within a continuous relaxation of the search space, enabling more efficient searching compared to discrete methods. 
Performance estimation strategies aim to cheaply estimate the quality (\eg, accuracy or efficiency) of an architecture, using techniques like proxy tasks and weight sharing to reduce the computational cost of NAS \cite{baker2017accelerating, falkner2018bohb}.
Proxy tasks involve training architectures on smaller or simplified versions of the target task to quickly evaluate their performance, while weight sharing trains a single super-network that contains all possible architectures within the search space, allowing for rapid evaluation without training each one from scratch.
All of these approaches typically search for optimal architectures using fully balanced datasets. However, our experiments demonstrate that the set of optimal architectures can vary significantly between balanced and imbalanced datasets. 
This finding underscores the need for developing new NAS methods or efficient adaptation strategies to search for optimal architectures on real-world, imbalanced datasets.

%In this section, we first introduce federated learning (FL), and then analyze several challenges in FL.
%Furthermore, we present the motivation to utilize multi-exit models in our system.
\subsection{Long-tailed data learning}
Class imbalance, particularly the long-tail distribution, is a significant challenge in many real-world applications. 
Long-tailed data refers to datasets where a few classes (head classes) have a large number of samples, while many other classes (tail classes) have relatively few samples.
Prior research on addressing long-tail imbalance can be broadly categorized into three primary approaches: (as detailed in the survey of \cite{zhang2023deep}). 
The first is class rebalancing, which aims to mitigate the effects of class imbalance by adjusting the training data or the loss function. 
It includes techniques such as data re-sampling \cite{he2008adasyn,chawla2002smote}, 
loss re-weighting \cite{kang2019decoupling, cui2019class, duggal2021har, duggal2020elf}, and logit adjustment \cite{menon2020long, tian2020posterior, zhang2021distribution}.
In data re-sampling, minority classes are oversampled or majority classes are undersampled. 
Loss re-weighting assigns higher weights to tail classes in the loss function and logit adjustment modifies the logits (outputs before the final activation function) in a way that accounts for class imbalance, thus helping the model to better distinguish between minority and majority classes.
The second is information augmentation which includes transfer learning \cite{wang2017learning, yin2019feature }, which leverages pre-trained models from balanced datasets, and data augmentation \cite{chu2020feature}, which generates additional samples for minority classes through techniques like GANs.
The third is Module Improvement, which focuses on enhancing the model's architecture and learning process. Module improvement encompasses techniques in representation learning \cite {liu2019large}, classifier design \cite{wu2020solving}, decoupled training \cite{kang2019decoupling}, and ensembling \cite{zhou2020bbn}.
Distinct from these existing approaches, the work in \cite{duggal2022imb} explores a novel direction for enhancing performance on long-tail datasets by optimizing the backbone architecture through neural architecture search. This new approach complements existing methods and can be used in conjunction with them to further improve accuracy and efficiency on imbalanced datasets.

\subsection{Architecture transfer}
Previous work on evaluates the robustness of architectures to distributional shifts in training datasets. 
Neural Architecture Transfer \cite{lu2021neural} investigates the transferability of architectures from large-scale to small-scale fine-grained datasets. 
However, this approach has two main limitations: it only considers balanced source and target datasets, and it assumes that all target datasets are known in advance, which is not practical for many industrial applications. 

NASTransfer \cite{panda2021nastransfer} addresses transferability between large-scale imbalanced datasets, including highly imbalanced datasets, like ImageNet-22k.
Their approach is practically useful for very large datasets (\eg, ImageNet-22k) for whom direct search is prohibitive, however when it is feasible (\eg, on ImageNet) direct search typically leads to better architectures than proxy search. 
Our work differs by focusing on directly adapting a super-network pre-trained on fully balanced datasets to imbalanced ones. This approach emphasizes efficiency by retraining only the linear classification head while keeping the backbone frozen. 
By doing so, we significantly reduce the computational effort required compared to performing a full search on the target dataset.
	% \vspace{-2mm}
	% \section{PRELIMINARIES}\label{sec:prelim}
        % \input{content/prelim.tex}

        \section{METHODOLOGY}\label{sec:design}
        
\begin{figure*}[!t]
	\centering
	\includegraphics[width=1.7\columnwidth]{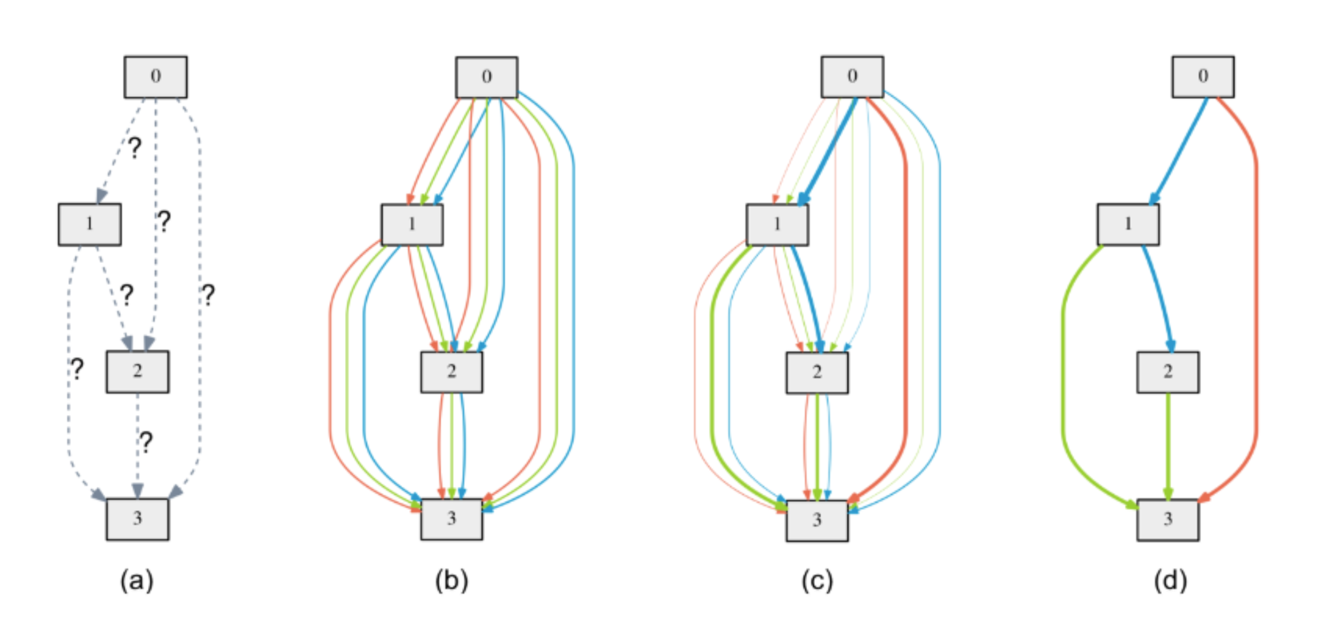}
	\caption{An overview of DARTS: (a) Operations on the edges are initially unspecified. (b) Continuous relaxation of the search space is achieved by placing a mixture of candidate operations on each edge. 
 (c) Joint optimization of the mixing probabilities and the network weights is performed by solving a bilevel optimization problem. (d) The final architecture is derived from the learned mixing probabilities.}
	\label{fig: DARTs}
	% \vspace{-0.5cm}
\end{figure*}
In this work, we mainly focus on searching for a super-network on the balanced dataset and adapting only the linear classifier on the target dataset to solve the computer vision problem (\eg, image classification).
We will introduce some technical details related to the scheme in \cite{duggal2022imb}, including the following three parts:
\subsection{Preliminaries}
Consider a training dataset $\mathcal{D}=\left\{x_1, y_i\right\}$, where $x_i$ denotes an image and $y_i$ its corresponding label.
Let $n_j$ be the number of training images in class $j$. 
Under the assumption of a long-tail distribution, after sorting classes by decreasing cardinality, we observe that $n_i \geq n_j$ for $i<j$, and $n_1>>n_C$. 
We denote a deep neural network by $\phi$, which comprises a backbone $\phi\left(\alpha, w_\alpha\right)$ with architecture $a$, weights $w_\alpha$ and a linear classifier $\phi\left(w_\theta\right)$. 
The model $\phi$ is trained using a combination of a training loss and a loss re-weighting strategy.

For balanced datasets, the network is typically trained using the standard cross-entropy loss (CE). In contrast, for imbalanced datasets, a re-weighting strategy  \cite{cui2019class} is applied to mitigate the bias towards majority classes. Specifically, samples from class $j$ are re-weighted by a factor of $\frac{1-\gamma}{1-\gamma^n j}$, where $\gamma$ is a hyperparameter controlling the degree of re-weighting. This technique helps to balance the influence of each class during training, thereby improving performance on minority classes.

The backbone $\phi\left(a, \mathbf{w}_\alpha\right)$ is responsible for feature extraction, transforming input images into high-level representations. These representations are then fed into the linear classifier $\phi\left(\mathbf{w}_\theta\right)$, which maps them to the final output classes. The training process involves optimizing both $\mathbf{w}_\alpha$ and $\mathbf{w}_\theta$ to minimize the overall loss.
To handle the class imbalance effectively, inspired by previous works \cite{duggal2020elf, cao2019learning}, the re-weighting strategy is often applied after a few initial epochs of standard training, known as delayed re-weighting (DRW).
This approach allows the model to first learn general features before focusing on underrepresented classes. The combined loss function can be expressed as:
\begin{equation}
    \mathcal{L}_{\text {total }}=\mathcal{L}_{\mathrm{CE}}+\lambda \mathcal{L}_{\mathrm{RW}}.
\end{equation}

where $\mathcal{L}_{\mathrm{CE}}$ is the cross-entropy loss, $\mathcal{L}_{\mathrm{RW}}$ is the re-weighted loss, and $\lambda$ is a scaling factor.

\subsection{Differentiable Architecture Search}

Differentiable Architecture Search (DARTS) represents a significant advancement in the field of neural architecture search (NAS). Traditional NAS methods often rely on reinforcement learning or evolutionary algorithms, which are computationally expensive due to the need to train and evaluate a large number of candidate architectures. In contrast, DARTS introduces a differentiable approach that allows for the efficient optimization of neural network architectures using gradient-based methods.

As shown in Fig. \ref{fig: DARTs}, in DARTS, the architecture search space is parameterized by a set of continuous variables that represent the probabilities of choosing different operations (\eg, convolutions, pooling) at each layer of the network. These continuous variables are optimized jointly with the network weights using standard gradient descent techniques. By formulating the search process as a differentiable problem, DARTS can efficiently explore the architecture space and converge to an optimal architecture in a fraction of the time required by traditional methods.

The DARTS framework consists of two main phases: the search phase and the evaluation phase. During the search phase, a super-network that encompasses all possible architectures within the search space is trained using a mixture of operations weighted by the learned continuous variables. 
The super-network is represented as a directed acyclic graph (DAG), where each node corresponds to a network layer, and each edge represents a candidate operation. 
We denote the super-network with backbone $\phi\left(\alpha, w_\alpha\right)$ and classifier $\phi\left(w_\theta\right)$ on a training dataset $D$ via the following minimization:
\begin{equation}
w_{\alpha, \mathcal{D}}^*, w_{\theta, \mathcal{D}}^*=\min _{w_\alpha, w_\theta} \underset{\alpha \sim \mathcal{A}}{\mathbb{E}}\left(\mathcal{L}\left(\phi\left(w_\theta\right), \phi\left(\alpha, w_\alpha\right) ; \mathcal{D}\right)\right).
\end{equation}
Here, $\mathcal{L}$ denotes the loss function and $\alpha \sim \mathcal{A}$ indicates sampling from the search space $\mathcal{A}$ via uniform, or
attentive sampling. 
The expectation $\mathbb{E}$ is computed over the sampled architectures $\alpha$, which are combined using the continuous variables.

The architecture of the neural network is represented as a weighted sum of candidate operations, where the weight of each operation is determined by the continuous variable $\alpha$. For each edge $(i, j)$ in the DAG, the operation $o^{(i, j)}$ is a weighted sum of all possible operations:
\begin{equation}
o^{(i, j)}(x)=\sum_{o \in \mathcal{O}} \frac{\exp \left(\alpha_o^{(i, j)}\right)}{\sum_{o^{\prime} \in \mathcal{O}} \exp \left(\alpha_{o^{\prime}}^{(i, j)}\right)} o(x)
\end{equation}

Here, $\mathcal{O}$ is the set of all candidate operations (e.g., convolutions, pooling), and $\alpha_o^{(i, j)}$ is the weight for operation $o$ on edge $(i, j)$. The softmax function ensures that the weights sum to 1, making the operation selection differentiable.

After the search phase, the learned continuous variables $\alpha$ are used to derive the final discrete architecture. 
The most likely operations, as indicated by $\alpha$, are selected to form the optimal architecture. This derived architecture is then retrained from scratch to validate its performance. The objective during this phase is to find the architecture $\alpha^*$ that maximizes the validation accuracy by follows:
% The second step involves searching the optimal architecture that maximizes validation accuracy via the following optimisation
\begin{equation}
\left.\alpha_{\mathcal{D}}^*=\max _{\alpha \in \mathcal{A}} \operatorname{Acc}\left(\phi\left(w_\theta\right), \phi\left(\alpha, w_\alpha\right) ; \mathcal{D}\right)\right.
\end{equation}

This maximization is commonly carried out using evolutionary algorithms or reinforcement learning techniques. In the following sections, we will explore efficient adaptation methods for modifying a NAS super-network, originally trained on a balanced dataset, to perform well on an imbalanced dataset.

\subsection{Rank adaptation procedures} 
Given source and target datasets $\mathcal{D}_s, \mathcal{D}_t$, the initial step involves training a train a super-network on $D_s$ by solving the following optimization problem:
\begin{equation}
\mathbf{w}^*_{\alpha,\mathcal{D}_s}, \mathbf{w}^*_{\theta,\mathcal{D}_s} = \min_{\mathbf{w}_\alpha, \mathbf{w}_\theta} \mathbb{E}_{\alpha \sim \mathcal{A}} \left( \mathcal{L} (\phi (\mathbf{w}_\theta), \phi (\alpha, \mathbf{w}_\alpha); \mathcal{D}_s) \right)
\end{equation}

The primary objective is to adapt the optimal super-network weight $w_{\alpha, \mathcal{D}_s}^*, w_{\theta, \mathcal{D}_s}^*$ obtained from the source dataset $D_s$ to the target dataset $D_t$ which is characterized by class imbalance. The most efficient strategy for this adaptation involves freezing the backbone of the network and adapting only the linear classifier on $D_t$ by minimizing the reweighted loss function $\mathcal{L}_{R W}$:
\begin{equation}
w_{\theta, \mathcal{D}_t}^*=\min _{w_\theta} \underset{\alpha \sim \mathcal{A}}{\mathbb{E}}\left(\mathcal{L}_{R W}\left(\phi\left(w_\theta\right), \phi\left(\alpha, w_{\alpha, \mathcal{D}_s}^*\right) ; \mathcal{D}_t\right)\right) .
\end{equation}
Here, $\mathcal{L}_{\mathrm{RW}}$ is the re-weighted loss function tailored to handle class imbalance. This method significantly reduces computational costs as only the classifier is retrained, while the backbone remains unchanged.
The super-network resulting from this procedure contains backbone weights $w_{\alpha, \mathcal{D}_s}^*$ trained on $\mathcal{D}_s$ and classifier weights $w_{\theta, \mathcal{D}_s}^*$ trained on $\mathcal{D}_t$. 
Alternatively, another approach involves fine-tuning both the backbone and the classifier on the target dataset. This procedure is more computationally intensive but allows for better adaptation to the new data distribution. 
This can be done by minimizing the delayed re-weighted loss $\mathcal{L}_{D R W}$:
\begin{equation}
w_{\alpha, \mathcal{D}_t}^{* *}, w_{\theta, \mathcal{D}_t}^*=\min _{w_\alpha, w_\theta} \underset{\alpha \sim \mathcal{A}}{\mathbb{E}}\left(\mathcal{L}_{D R W}\left(\phi\left(w_\theta\right), \phi\left(\alpha, w_{\alpha, \mathcal{D}_s}^*\right) ; \mathcal{D}_t\right)\right) \text {. }
\end{equation}

In this context, the double star on $w_{\alpha, \mathcal{D}_t}^{* *}$ indicates the weights were obtained via fine-tuning $w_{\alpha, \mathcal{D}_s}^*$ using a reduced learning rate and fewer training epochs.
It's important to note that the delayed re-weighted loss $\mathcal{L}_{D R W}$ starts as the unweighted loss $\mathcal{L}$ during the initial epochs and transitions to the re-weighted loss $\mathcal{L}_{R W}$ in later epochs. Although this second adaptation procedure is more computationally demanding because it also involves updating the backbone, it is still much less intensive than performing a full search on the target dataset.

The third and most computationally demanding method involves conducting a direct search on the target dataset using the re-weighted loss $\mathcal{L}_{D R W}$. This approach aims to identify the optimal architecture and weights from scratch, using the following optimization:
\begin{equation}
w_{\alpha, \mathcal{D}_t}^*, w_{\theta, \mathcal{D}_t}^*=\min _{w_\alpha, w_\theta} \underset{\alpha \sim \mathcal{A}}{\mathbb{E}}\left(\mathcal{L}_{D R W}\left(\phi\left(w_\theta\right), \phi\left(\alpha, w_\alpha\right) ; \mathcal{D}_t\right)\right)
\end{equation}

Although this method incurs high computational costs, it potentially yields the most tailored architecture for the imbalanced dataset.
The three adaptation strategies are summarized in Table \ref{tab: summary}.

\begin{figure*}[t]
\centering
\subfigure[Balance]{
\includegraphics[width=0.22\textwidth,height=3.9cm]{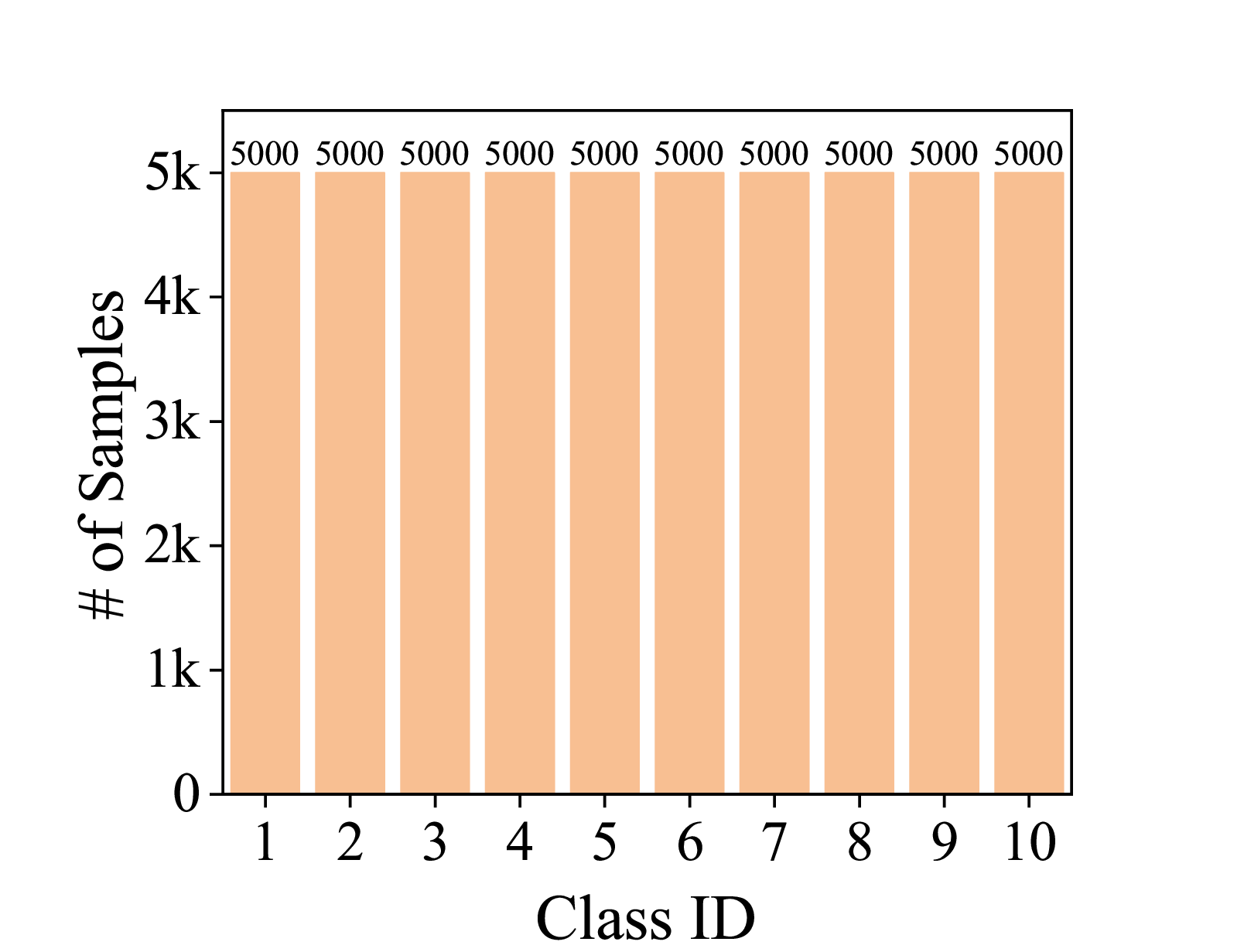}\label{fig: balance}}
\subfigure[Exponential($u$=0.1)]{
    \includegraphics[width=0.23\textwidth,height=3.9cm]{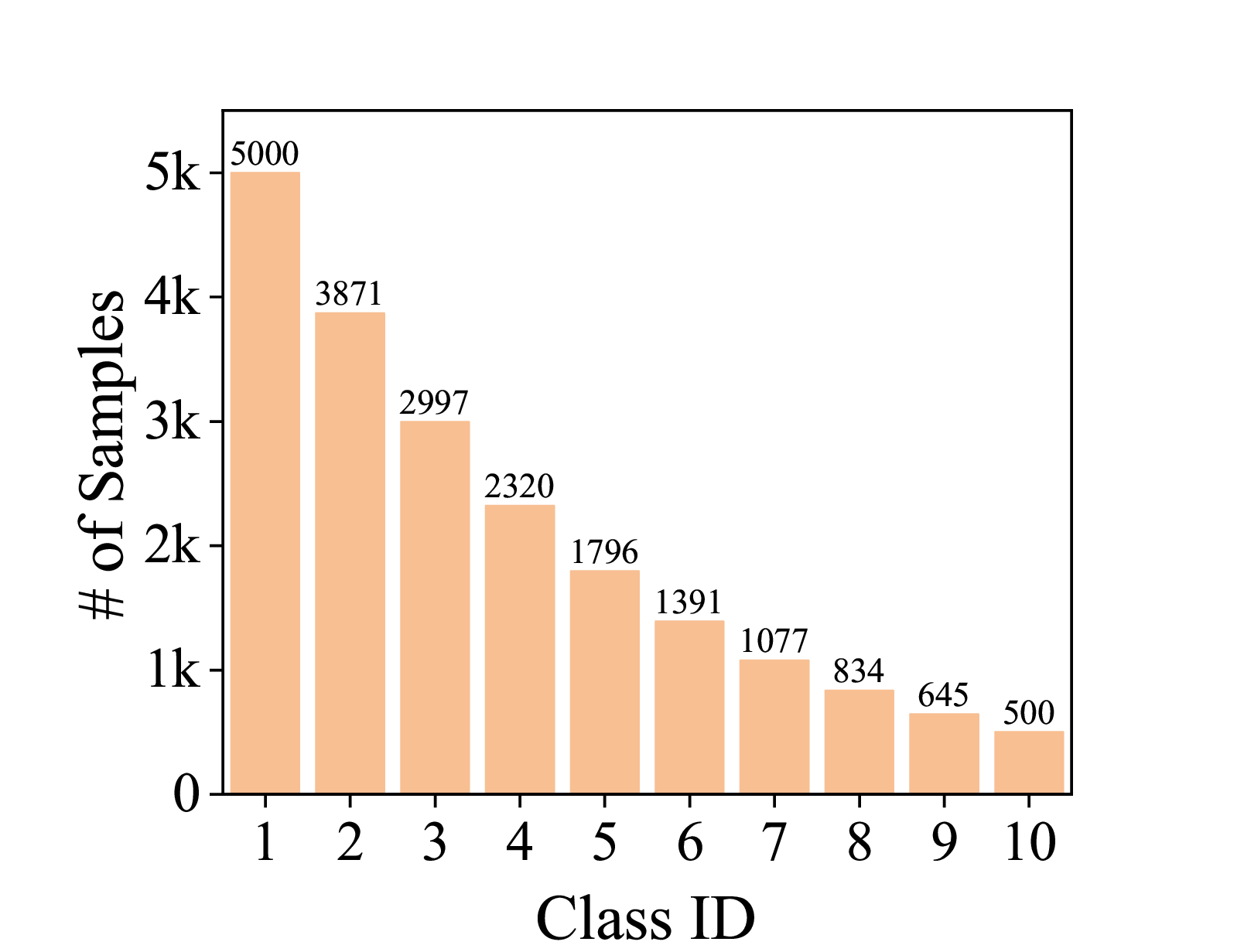}\label{fig: CIFAR10_waiting_time}}
\hspace{0.4cm}
\subfigure[Exponential($u$=0.01)]{
    \includegraphics[width=0.23\textwidth,height=3.9cm]{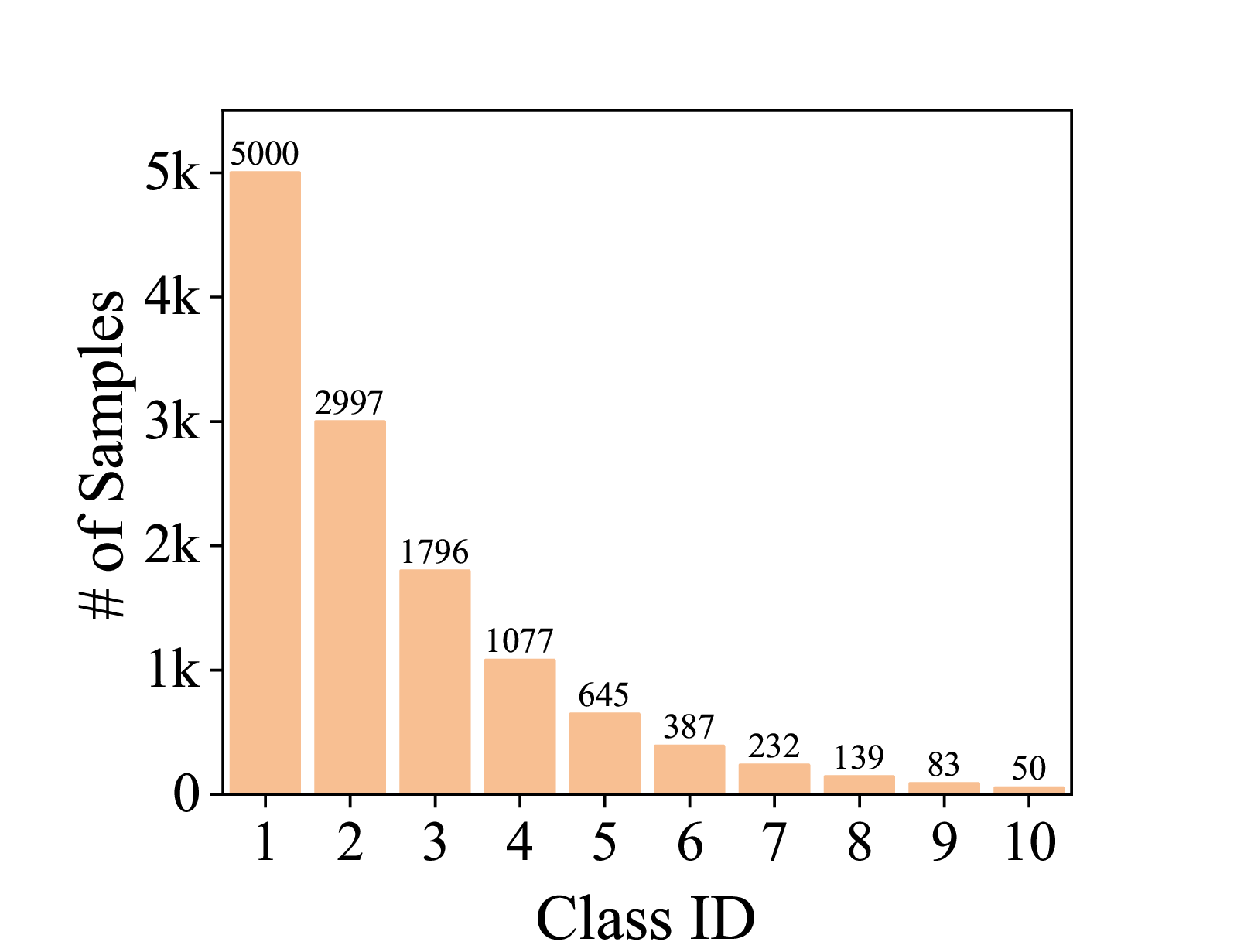}\label{fig: IMAGE_waiting_time}}
\subfigure[Step($u$=0.01)]{
    \includegraphics[width=0.23\textwidth,height=3.9cm]
    {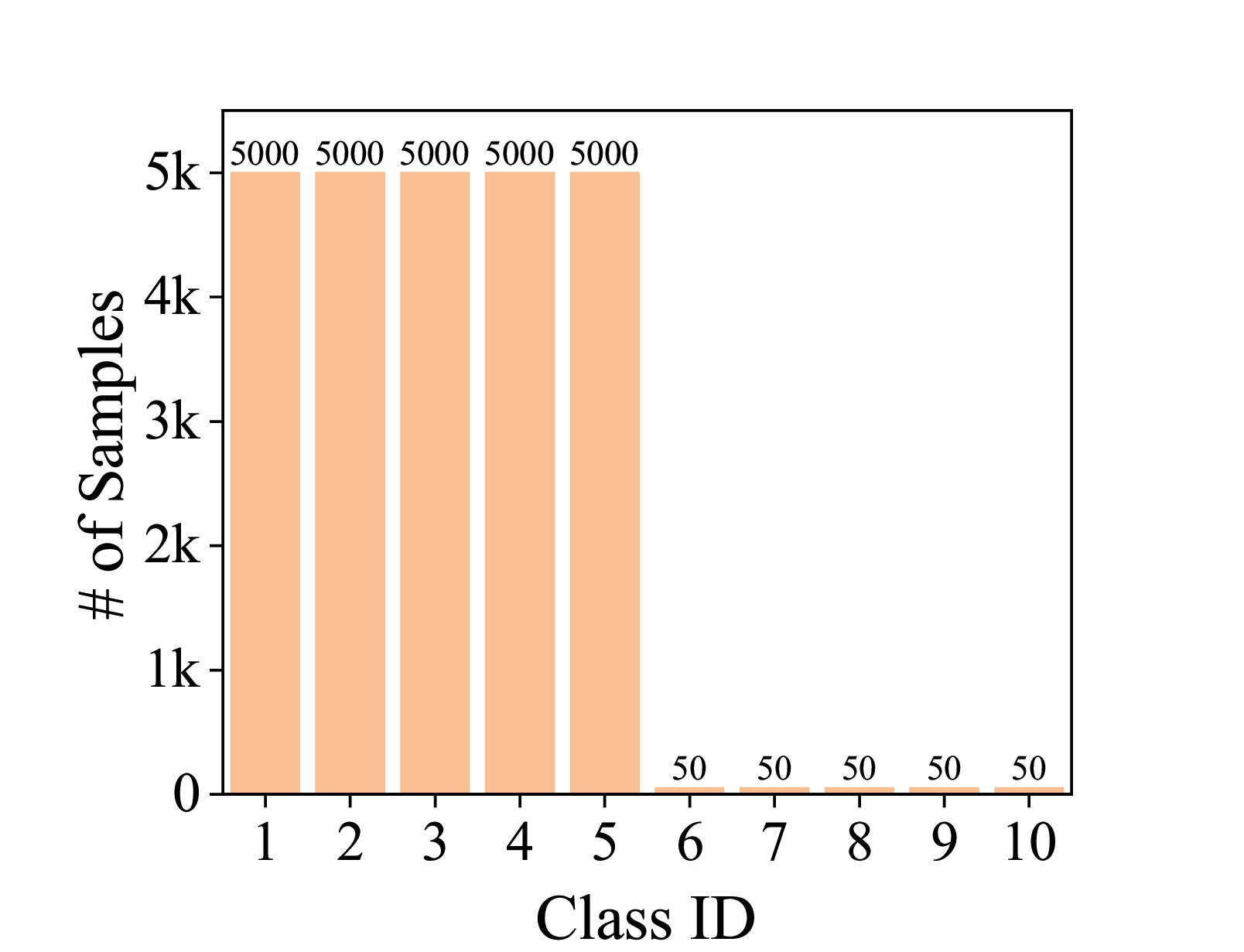}
    }
\caption{Four Different Label Distributions of CIFAR-10 dataset.} \label{fig: data distributions}
% \vspace{-0.3cm}
\end{figure*}
% \begin{\alpharray}{ccl}
% \hline \text { Adj } & \text { Eqn } & \text { Description } \\
% \hline \text { P0 } & \sqrt{3} & \text { No adaptation. } \\
% \text { P1 } & \sqrt{4} & \text { Freeze backbone, retrain classifier on } \mathcal{D}_t . \\
% \text { P2 } & \sqrt[5]{5} & \text { Finetune backbone and retrain classifier on } \mathcal{D}_t \\
% \text { P3 } & 6 & \text { Re-train backbone and classifier on } \mathcal{D}_t . \\
% \hline
% \end{array}

\begin{table}[t]
	\centering
    \vspace{-0.2cm}
	\caption{Summary of rank adaptation procedures.}\label{tab: summary}
	\begin{threeparttable}
		%		\resizebox{0.92\textwidth}{!}{%
		\resizebox{0.45\textwidth}{!}{%
			\begin{tabular}{ccc}
				\hline
				% \ & \multicolumn{8}{c}{The confidence threshold $C$}  \\
                \text { Symbol } & \text { Eqn } & \text { Details } \\
				\hline
				\text { P0 } & (5) & \text { No adaptation. } \\
				\hline
				\text { P1 } & (6) & \text { Freeze backbone, retrain classifier on $\mathcal{D}_t$.}  \\
                \hline
				% %				\hline
				\text { P2 } & (7) & \text { Finetune backbone and retrain classifier on $\mathcal{D}_t$.} 
                  \\
				\hline
                \text { P2 } & (8) & \text {  Re-train backbone and classifier on $D_t$.}  \\
				\hline
			\end{tabular}
		}
	\end{threeparttable}
    \vspace{-0.2cm}
\end{table}

        \section{Experiment Evaluation}\label{sec:evaluation}

% The evaluation focuses on three primary aspects: the effectiveness of different rank adaptation procedures, detailed ablation studies to understand various influencing factors, and overall performance comparison in terms of computational efficiency and accuracy.
\subsection{Implementation details}

% Furthermore, communication channels among clients and between the clients and the PS are established based on MPI (Message Passing Interface) \cite{gropp1999using}, which provides a set of sending and receiving functions to achieve efficient parallel communication.
% To reflect the heterogeneity and dynamics of network conditions, we fluctuate the inbound bandwidth of each client between 5Mb/s and 15Mb/s. 
% While the outbound bandwidth is typically smaller than the inbound bandwidth in a typical WAN \cite{wang2022accelerating}, we configure it to fluctuate within the range of [1.6Mb/s, 5Mb/s]. 
% As for computation heterogeneity, the time taken for one iteration on a client follows the Gaussian distribution whose 
% mean and variance are determined based on the test records on commercial devices (\eg, NVIDIA Jetson TX2, Xavier NX).

\textbf{Datasets.} To simulate real-world class imbalance scenarios, we constructed imbalanced versions of the CIFAR-10 and CIFAR-100 datasets by sub-sampling from their original training splits. 
Concretely, CIFAR-10 contains a total of 60,000 32$\times$32 color images, which are classified into 10 distinct classes. The dataset is divided into two subsets for training and testing purposes,
with 50,000 for training and 10,000 for testing. 
The images in CIFAR-10 are 32$\times$32$\times$3 dimensional.
CIFAR-100 has the same number of images as CIFAR-10 but consists of 100 classes, which makes it more challenging to train models for classification.
With the CIFAR-10-LT dataset and CIFAR-100-LT dataset, we explored three types of imbalance following \cite{cao2019learning}:
\begin{itemize}
\item \textbf{Balance:} Every class contains all the samples from the original dataset, \ie, 5,000.
\item \textbf{Step:} The latter half of the classes have the number of samples adjusted by a specific factor, resulting in each class having 5000 * factor samples. 
\item \textbf{Exponential:} For each class $i \in\{0,1, \ldots, 9\}$, the number of samples is set to $5000 \times$ factor $^{(i / 9)}$. 
\end{itemize}
 % ese settings allow us to systematically study the impact of different imbalance types on model performance.

Specifically, our experiments involve four different label distributions. 
Taking CIFAR-10 for example, the sample distribution of CIFAR-10 is illustrated in Fig. \ref{fig: data distributions}. By Figs. \ref{fig: data distributions}(b)-\ref{fig: data distributions}(c), it can be found that the smaller the imbalance factor $\mu$, the more pronounced the long tail characteristics of the label distribution.

\textbf{Sub-network and super-network training details.} Following IMB-NAS, we trained a network on balanced CIFAR-10 and CIFAR-100 datasets for 200 epochs. The training process began with an initial learning rate of 0.1, which was decayed by a factor of 0.01 at epochs 160 and 180. 
The cross-entropy loss function was used to optimize the model. For the imbalanced versions of these datasets, an effective re-weighting strategy inspired by \cite{cui2019class} is implemented.

We trained a super-network for 500 epochs, starting with an initial learning rate of 0.1. The learning rate was reduced by a factor of 0.01 at epochs 300 and 400. 
For imbalanced datasets, re-weighting is applied at epoch 350 to address class imbalance issues.
To identify the best subnet, we employed an evolutionary search strategy as described \cite{guo2020single}. 
This involved 20 generations with a population of 50, utilizing a crossover number of 25, mutation number of 25, a mutation probability of 0.1, and selecting the top-k of 10 for the final architecture selection.

\textbf{Adaptation Strategies.}
To adapt a super-network from a balanced to an imbalanced dataset, as described in \cite{duggal2022imb}, we fine-tuned the network for 200 epochs with an initial learning rate of 0.01, which was decayed by a factor of 0.01 at epoch 100. 
For procedure $P 1$, we introduce re-weighting at epoch 1. For P2, we delay the re-weighting to epoch 100. For P3, we follow the NAS strategy detailed above.

\textbf{Experiment Setup.} We performed our experiments using an AMAX deep learning workstation. This setup is equipped with an Intel(R) Xeon(R) Gold 5218R CPU, 8 NVIDIA GeForce RTX 3090 GPUs, and 256 GB RAM. All model training and evaluation were implemented using the PyTorch framework \cite{paszke2019pytorch}, ensuring a robust and scalable environment for our extensive simulations.

\subsection{IMB-NAS Evaluation} 
\textbf{ }
Given a NAS super-network trained on a source dataset $D_s$, the goal is to adapt it efficiently to a target dataset $D_t$. 
% Subsequently, the best sub-network is searched within the adapted super-network. 
Table \ref{tab:my_label} presents the results for scenarios where $D_s$ is CIFAR-10 and $D_t$ is CIFAR-100 with varying levels of imbalance.
The baseline (P0) involves retraining the best sub-networks from $D_s$ to $D_t$.
The optimal scenario (P3) entails direct training on $D_t$, representing the highest accuracy achievable.

The adaptation procedures (P1 and P2 ) are displayed in the middle rows.  
Both adaptation methods surpass the baseline at higher imbalance levels, indicating that architectures optimized on $D_s$ do not transfer well to imbalanced $D_t$.
Surprisingly, we obtained similar results as IMB-NAS that
$P1$ consistently outperforms P2. This result is unexpected since P2 involves adapting the NAS backbone with the target data, whereas P1 retains the backbone from the source dataset. 
This suggests that class imbalance poses a greater challenge for NAS backbone optimization than the domain differences between CIFAR-10 and CIFAR-100.
Also we find that P1 and P2 achieve accuracy levels close to the P3 while avoiding much of the computational cost associated with P3.
This efficiency is crucial for practical applications where computational resources are limited. 
These findings underscore the importance of selective adaptation strategies in handling imbalanced datasets, highlighting the necessity for tailored approaches in NAS applications.

% As observed from Tables \ref{tab:my_label}, both adaptation procedures significantly outperform the baseline at higher levels of imbalance. 
% This indicates that the architectures optimized on $D_s$ cannot be assumed to be optimal for imbalanced target datasets. 
% Notably, between P1 and P2, P1 consistently delivers better performance. 
% This is unexpected since P2 also adapts the NAS backbone using the target data, whereas P1 retains the backbone from the source dataset. 
% We hypothesize that this is because class imbalance poses a greater challenge for NAS backbone optimization than the domain differences between CIFAR-10 and CIFAR-100.
\begin{table}
    \centering
    \scalebox{0.8}{
    \begin{tabular}{cccccc}
        \hline & Symbol & \multicolumn{4}{c}{ Imbalance Ratio (factor) } \\
        \cline { 3 - 6 } & & $0(Balance) $ & $0.1(Exp)$  &  $0.01(Exp)$ &  $0.01(Step)$ \\
        \hline baseline & P0 & 52.83 & 48.43 & 41.18 &  38.43\\
        \hline & P1 & $52.76$ & $49.39$  & $42.37$ &  39.58\\
        & P2 & 52.71 & 49.05&  42.02 & 39.21 \\
        \hline paragon & P3 & 52.86 & 49.54 & 42.32 & 39.49 \\
        \hline
    \end{tabular}}
    \caption{\text{CIFAR10-Balanced} $\longrightarrow$ \text {CIFAR100}}
    \label{tab:my_label}
\end{table}

	\section{Conclusions} \label{sec:conclusion}
	This work aims to improve performance on class-imbalanced datasets by optimizing the backbone architecture.  
We begin by reviewing related works on NAS and various techniques for handling long-tailed datasets, providing a thorough understanding of the current landscape.
Our investigation revealed that an architecture's performance on balanced datasets does not consistently predict its effectiveness on imbalanced datasets. 
This insight implies that re-running NAS for each target dataset might be necessary. To avoid the substantial computational cost of re-running NAS,
we explored an existing approach called IMB-NAS.
This innovative method proposes adapting a NAS super-network trained on balanced datasets to imbalanced ones.
IMB-NAS introduces several adaptation methods and discovers that re-training the linear classification head while keeping the NAS super-network backbone frozen outperforms other adaptation strategies.
To further understand IMB-NAS, we conducted a series of experiments on the long-tailed dataset to evaluate its performance.
Our experimental results generally aligned with our expectations, confirming the effectiveness of IMB-NAS.

	% \IFCLASSOPTIONcompsoc
	%   % The Computer Society usually uses the plural form
	%   \section*{Acknowledgments}
	% \else
	%   % regular IEEE prefers the singular form
	%   \section*{Acknowledgment}
	% \fi
	% \input{content/ack.tex}
	
	%\vspace{-0.3em}
	%\section{Acknowledgement}\label{sec:Acknowledgement}
	%\input{content/ack.tex}
	
	%\section*{Acknowledgement}
	%This search of Xu, Yu and Liu is supported by NSFC with No. U1301256 and 61472383, NSF of Anhui in China under No. 1408085MKL08, and NSF of Jiangsu in China under No. BK20161257. The research of Qian is supported by UC Santa Cruz Startup Grant and NSF grant CNS-1464335. The research of Li is partially supported by NSFC with No. 61520106007, China National Funds for Distinguished Young Scientists with No. 61625205, Key Research Program of Frontier Sciences, CAS, No. QYZDY-SSW-JSC002, NSF ECCS-1247944, NSF CMMI 1436786, and NSF CNS 1526638.
	
	%%\vspace{-1em}
	\bibliographystyle{IEEEtran}
	\bibliography{content/NAS.bib}

\end{document}